\documentclass[letterpaper, 10 pt, conference]{ieeeconf}
\IEEEoverridecommandlockouts
\let\labelindent\relax
\usepackage{graphicx} \graphicspath{ {figures/} }
\usepackage{amsmath,amssymb,mathabx}
\usepackage{acronym}
\usepackage{enumitem}
\usepackage{booktabs}
\usepackage{hyperref}
\usepackage{balance}
\usepackage{xspace,setspace}
\usepackage[skip=3pt,font=small]{subcaption}
\usepackage[skip=3pt,font=small]{caption}
\usepackage[dvipsnames]{xcolor}
\usepackage[capitalise]{cleveref}
\usepackage{tabularx,colortbl,multirow,array,makecell}
\usepackage{overpic}
\usepackage{cite}

\makeatletter
\DeclareRobustCommand\onedot{\futurelet\@let@token\@onedot}
\def\@onedot{\ifx\@let@token.\else.\null\fi\xspace}
\def\eg{\emph{e.g}\onedot} 
\def\ie{\emph{i.e}\onedot} 
 
\def\etc{\emph{etc}\onedot} 
\def\wrt{w.r.t\onedot} 
\def\etal{\emph{et al}\onedot}
\makeatother

\crefname{algocf}{alg.}{algs.}
\Crefname{algocf}{Algorithm}{Algorithms}

\makeatletter
\def\BState{\State\hskip-\ALG@thistlm}
\makeatother

\makeatletter
\renewcommand{\paragraph}{%
  \@startsection{paragraph}{4}%
  {\z@}{0ex \@plus 0ex \@minus 0ex}{-1em}%
  {\hskip\parindent\normalfont\normalsize\bfseries}%
}
\makeatother

\crefname{algocf}{alg.}{algs.}
\Crefname{algocf}{Algorithm}{Algorithms}

\definecolor{gblue}{HTML}{4285F4}
\definecolor{gred}{HTML}{DB4437}

\acrodef{dof}[DoF]{Degree of Freedom}
\acrodef{vkc}[VKC]{Virtual Kinematic Chain}
\acrodef{tamp}[TAMP]{Task and Motion Planning}
\acrodef{pddl}[PDDL]{Planning Domain Definition Language}
\acrodef{rrt}[RRT]{Rapidly-exploring Random Tree}
\acrodef{ompl}[OMPL]{Open Motion Planning Library}
\acrodef{iws}[IWS]{Iterated Width Search}
\acrodef{bfs}[BFS]{Breadth First Search}

\title{\LARGE \bf Efficient Task Planning for Mobile Manipulation:\\a Virtual Kinematic Chain Perspective}

\author{Ziyuan Jiao$^{1*}$\quad{}Zeyu Zhang$^{1*}$\quad{}Weiqi Wang$^{1}$\quad{}David Han$^{2}$\quad{}Song-Chun Zhu$^{1}$\quad{}Yixin Zhu$^{1}$\quad{}Hangxin Liu$^{1}$
    \thanks{$^{*}$ Ziyuan Jiao and Zeyu Zhang contributed equally to this work.}
    \thanks{$^{1}$ UCLA Center for Vision, Cognition, Learning, and Autonomy (VCLA) at Statistics Department. Emails:
    {\fontsize{7.5}{8}\selectfont\tt\{zyjiao, zeyuzhang, weiqi.wang, yixin.zhu, hx.liu\}@ucla.edu},
    \tt{sczhu@stat.ucla.edu}.}%
    \thanks{$^{2}$ Drexel University, Department of Electrical and Computer Engineering.
Email:
    \tt{dkh42@drexel.edu}.}%
    \thanks{The work reported herein was supported by ONR N00014-19-1-2153, ONR MURI N00014-16-1-2007, and DARPA XAI N66001-17-2-4029.}%
}

\begin{document}

\maketitle
\thispagestyle{empty}
\pagestyle{empty}

\begin{abstract}
We present a \ac{vkc} perspective, a simple yet effective method, to improve task planning efficacy for mobile manipulation. By consolidating the kinematics of the mobile base, the arm, and the object being manipulated collectively as a whole, this novel \ac{vkc} perspective naturally defines \textit{abstract actions} and eliminates unnecessary predicates in describing intermediate poses. As a result, these advantages simplify the design of the planning domain and significantly reduce the search space and branching factors in solving planning problems. In experiments, we implement a task planner using \ac{pddl} with \ac{vkc}. Compared with conventional domain definition, our \ac{vkc}-based domain definition is more efficient in both planning time and memory. In addition, abstract actions perform better in producing feasible motion plans and trajectories. We further scale up the \ac{vkc}-based task planner in complex mobile manipulation tasks. Taken together, these results demonstrate that task planning using \ac{vkc} for mobile manipulation is not only natural and effective but also introduces new capabilities.
\end{abstract}

\section{Introduction}

As one of the central themes in AI and robotics, task planning is typically solved by searching a feasible action sequence in a domain. Researchers have demonstrated a wide range of successful robotics applications~\cite{lavalle2006planning,karpas2020automated} with effective representations or programming languages, such as STRIPS~\cite{fikes1971strips}, hierarchical task network~\cite{nau2003shop2}, temporal and-or-graph~\cite{edmonds2019tale,liu2019mirroring}, Markov decision process~\cite{bellman1957markovian}, and \ac{pddl}~\cite{fox2003pddl2}.

An effective task planner in robotics generally possesses two characteristics. First, the planning domain must be clearly designed, which includes a set of predicates that truthfully describe the environment states, a set of actions that specify how states transit, and a goal specification that indicates the desired result. However, the definitions of these components are tightly coupled; thus, designing the planning domain could be tedious and error-prone. Second, the abstract notion of symbolic actions should be realizable by motion planners; \ie, the design of these abstract symbols should have practical meaning. These two requirements pose additional challenges in task planning for mobile manipulation; the robot consists of a mobile base and an arm, which possess different motion patterns and capabilities.

\begin{figure}[t!]
    \vspace{-3pt}
    \centering
    \includegraphics[width=\linewidth]{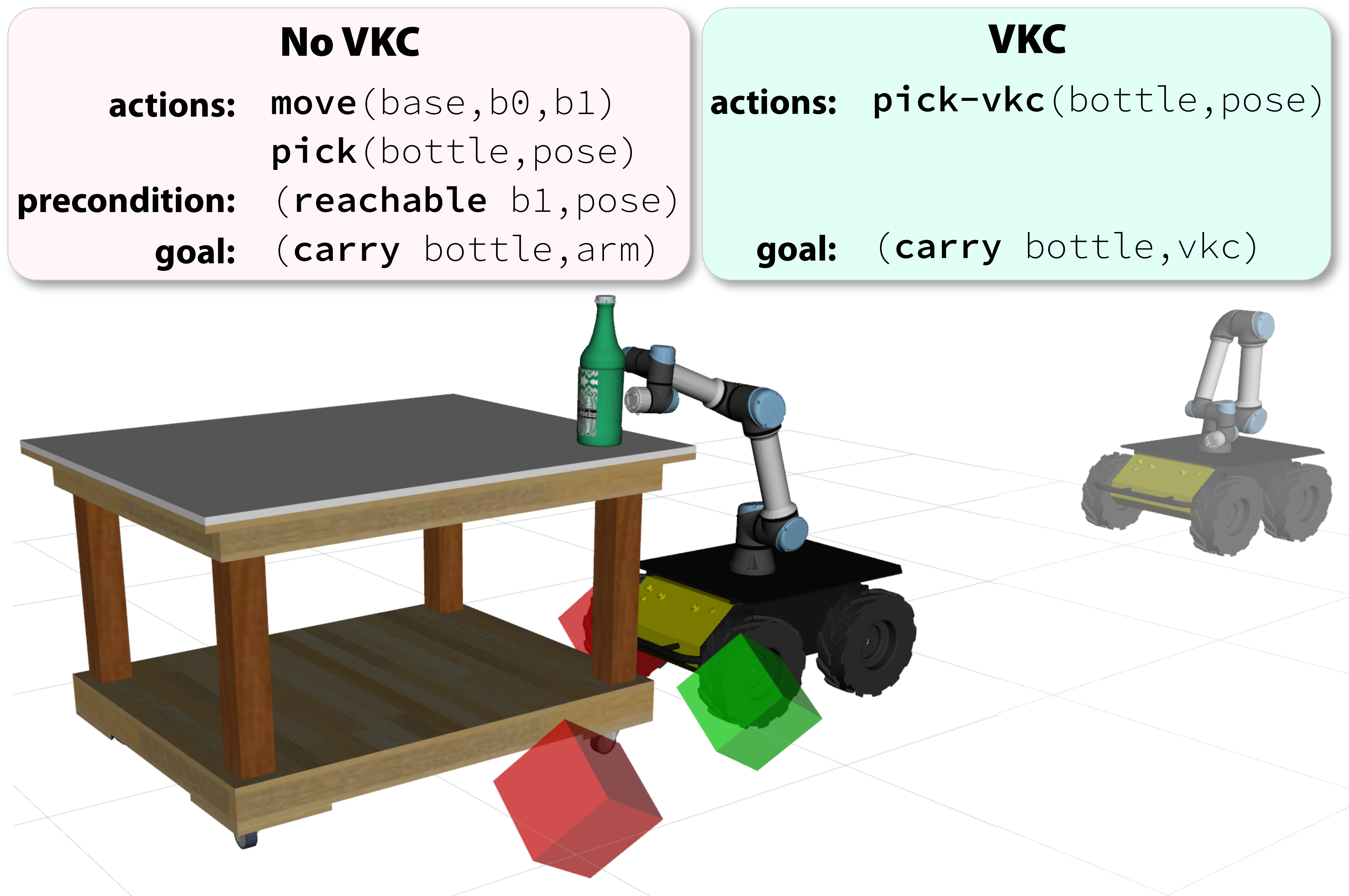}
    \caption{\textbf{A typical task planning setup, wherein the mobile manipulator is to tasked to navigate and pick up the object on the desk.} The \ac{vkc}-based domain specification reduces the search space by removing the poses of the mobile base near red cubes, resulting in a simpler and more intuitive task planning domain.}
    \label{fig:motivation}
\end{figure}

To clearly illustrate the above challenges, let us take \cref{fig:motivation} as a concrete example, wherein a mobile manipulator is tasked to navigate and pick up the bottle on the desk. A dedicated set of predicates and actions must be specified for the mobile base and the arm; for instance, moving the \textit{base} (\texttt{move($\cdot$)}) to a configuration, such that the \textit{arm} can pick up the object (\texttt{pick($\cdot$)}). Of note, finding such a pose oftentimes requires to specify the mobile base and the arm \textit{individually}. However, this separation in the planning domain is \textit{artificial} in nature and ineffectively introduces an \textit{unnecessarily} larger planning space: The valid poses of the mobile base near the goal (\ie, the bottle) must be specified (indicated by the cubes in \cref{fig:motivation}) in advance, and exactly one (\eg, the green cube) must be selected via sampling or searching under pre-defined heuristic or criteria. This deficiency becomes increasingly evident as the task sequence grows longer and prohibits natural motions that require foot-arm coordination; coordinating the base and arm movements remains challenging even for existing whole-body motion planning methods~\cite{shankar2016kinematics,bodily2017motion,chitta2010planning}, let alone realizing a symbolic task plan with a feasible motion plan.

In this paper, we seek to tackle these challenges in task planning with a focus on domain specification. Specifically, we ask: (i) Is there an intermediate representation to facilitate the specification of the planning domain such that the search can be more natural and efficient, without relying on manually-designed heuristics? (ii) Can this new representation simplify the planning domain in long-horizon tasks while preserving motion feasibility, especially for those who require complex foot-arm coordination?

\setstretch{0.97}

In particular, we propose a \acf{vkc} perspective for mobile manipulation, which consolidates the kinematics of the mobile base, the arm, and the object being manipulated into a single kinematic model. By treating the robot as a whole, more abstract actions can be defined to jointly account for both the base and the arm; see \texttt{pick-vkc($\cdot$)} \textit{vs} \texttt{move($\cdot$)} and \texttt{pick($\cdot$)} in \cref{fig:motivation}. Such an abstraction alleviates the manually-defined \textit{heuristic} of where the robot can reach the goal and the unnecessary definitions of \textit{intermediate goals}, \eg, predicates describing the robot's pose before reaching the goal. As a result, this modification of the planning domain reduces the branching factor, making it scalable to more complex tasks. Crucially, the abstraction introduced by \ac{vkc} does not sacrifice the success rate to generate a solvable motion planning problem.

In experiments, we implement the \ac{vkc}-based task planning using \ac{pddl}. Compared with a standard \ac{pddl} implementation, the \ac{vkc} perspective simplifies the domain setup and exhibits significant improvements in planning time and memory. Moreover, we demonstrate that abstract actions introduced by \ac{vkc} are executable at the motion level; they can generate solvable motion planning problems in the form of trajectory optimization or sampling methods. We further validate \ac{vkc}-based task planning in long-horizon tasks requiring foot-arm coordination. Taking together, our \ac{vkc} perspective offers a simple yet effective intermediate representation for domain specification in task planning.

\subsection{Related Work}

\paragraph*{\ac{tamp} in mobile manipulation}

Thanks to the development of \ac{pddl} and other planning architectures, complex symbolic task planning can be solved using standard algorithms~\cite{karpas2020automated}. Hence, the community has shifted the focus to corresponding a valid symbolic action sequence to feasible motions, which leads to the field of \ac{tamp}~\cite{garrett2020integrated}. While researchers tackle this problem from various angles, such as incorporating motion-level constraints to the task planning~\cite{erdem2011combining,kaelbling2011hierarchical,garrett2018ffrob}, developing interfaces that communicate between task and motion~\cite{srivastava2014combined}, or inducing abstracted modes from motions~\cite{toussaint2015logic,toussaint2018differentiable}, it remains a largely unsolved problem. In addition, movements of a mobile base and a manipulator are commanded by two or more separate actions~\cite{bidot2017geometric,garrett2018ffrob,kim2019learning}, causing increased planning time, less coordinated movements, \etc. In comparison, the \ac{vkc} perspective serves as an intermediate representation that benefits the task modeling of mobile manipulation, improves computation efficacy, and facilitates motion planning.

\paragraph*{\ac{vkc} in robot modeling and planning}

The idea of \ac{vkc} could be traced back to 1997 by Pratt~\etal~\cite{pratt1997virtual}, who proposed \ac{vkc} to design bipedal robot locomotion~\cite{pratt2001virtual}. Later, this idea was adopted to chain serial manipulators to form one kinematic chain~\cite{likar2014virtual} and to dual-arm manipulation tasks; for instance, connecting parallel structures via rigid-body objects~\cite{wang2015cooperative}, and modeling whole-body control of mobile manipulators~\cite{wang2016whole}. Recently, \ac{vkc} is also adopted for wheeled-legged robot control~\cite{laurenzi2019augmented}. In this paper, we further push the idea of \ac{vkc} and demonstrate its advantages as an intermediate representation in modeling and planning complex mobile manipulation tasks.

\setstretch{1}

\subsection{Overview}

The remainder of this paper is organized as follows. \cref{sec:modeling} introduces the modeling process of constructing a \ac{vkc}. A tasking planning framework from the \ac{vkc} perspective based on \ac{pddl} is described in \cref{sec:task}. \cref{sec:motion} further illustrates how the high-level \ac{vkc}-based task planning facilitates low-level motion planning. In a series of mobile manipulation tasks, we demonstrate the efficacy of \ac{vkc}s with a high success rate in \cref{sec:exp}. We conclude the paper with discussions in \cref{sec:conclusion}.

\begin{figure}[t!]
    \centering
    \begin{subfigure}[b]{0.5\linewidth}
        \centering
        \includegraphics[width=\linewidth,trim={0cm 0.7cm 0cm 0cm},clip]{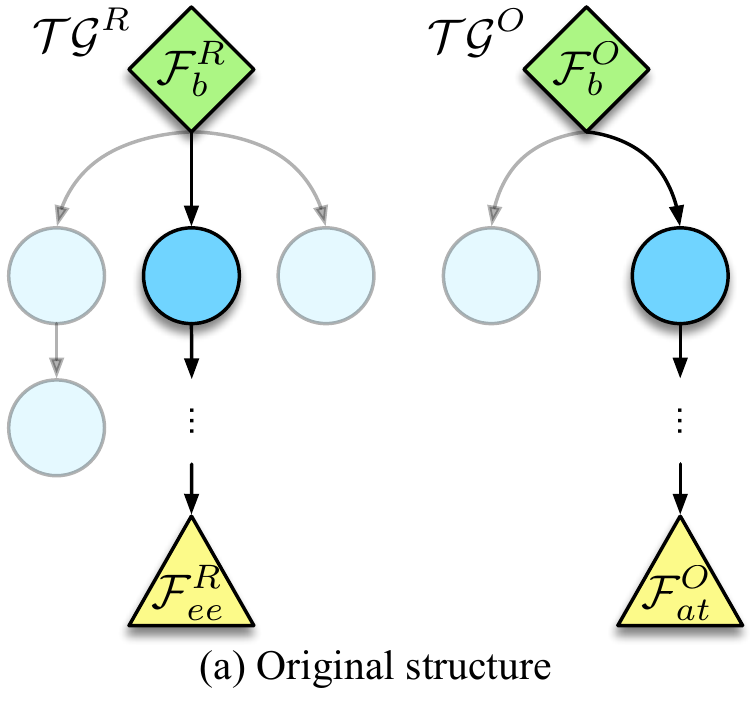}
        \caption{Original Structure}
        \label{fig:vkc_construction_a}
    \end{subfigure}%
    \begin{subfigure}[b]{0.5\linewidth}
        \centering
        \includegraphics[width=\linewidth,trim={0cm 0.7cm 0cm 0cm},clip]{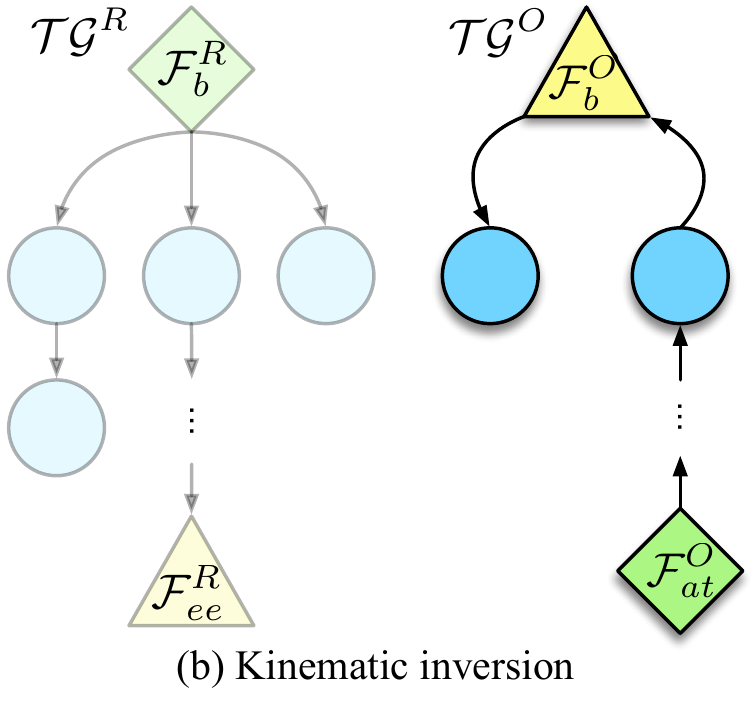}
        \caption{Kinematic Inversion}
        \label{fig:vkc_construction_b}
    \end{subfigure}%
    \\
    \begin{subfigure}[b]{0.5\linewidth}
        \centering
        \includegraphics[width=\linewidth,trim={0cm 0cm 0cm 0cm},clip]{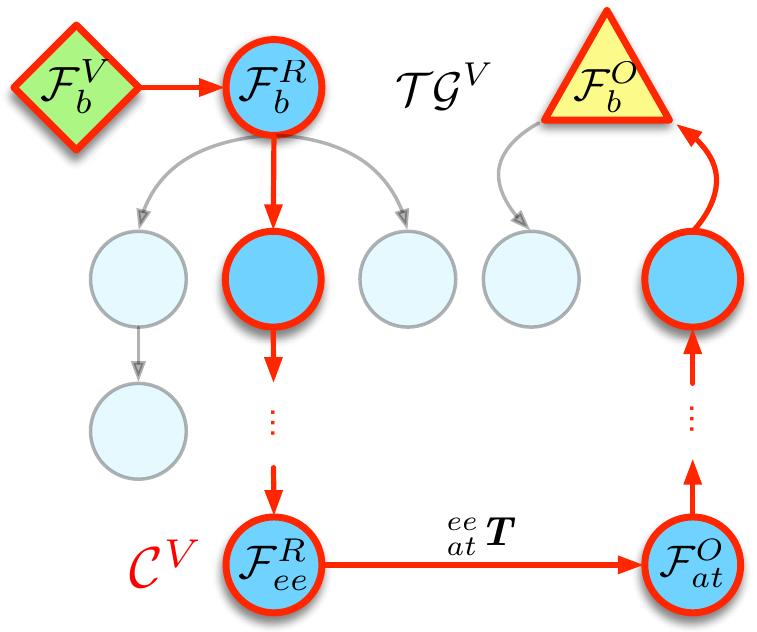}
        \caption{\ac{vkc} Construction}
        \label{fig:vkc_construction_d}
    \end{subfigure}%
    \begin{subfigure}[b]{0.5\linewidth}
        \centering
        \includegraphics[width=0.85\linewidth]{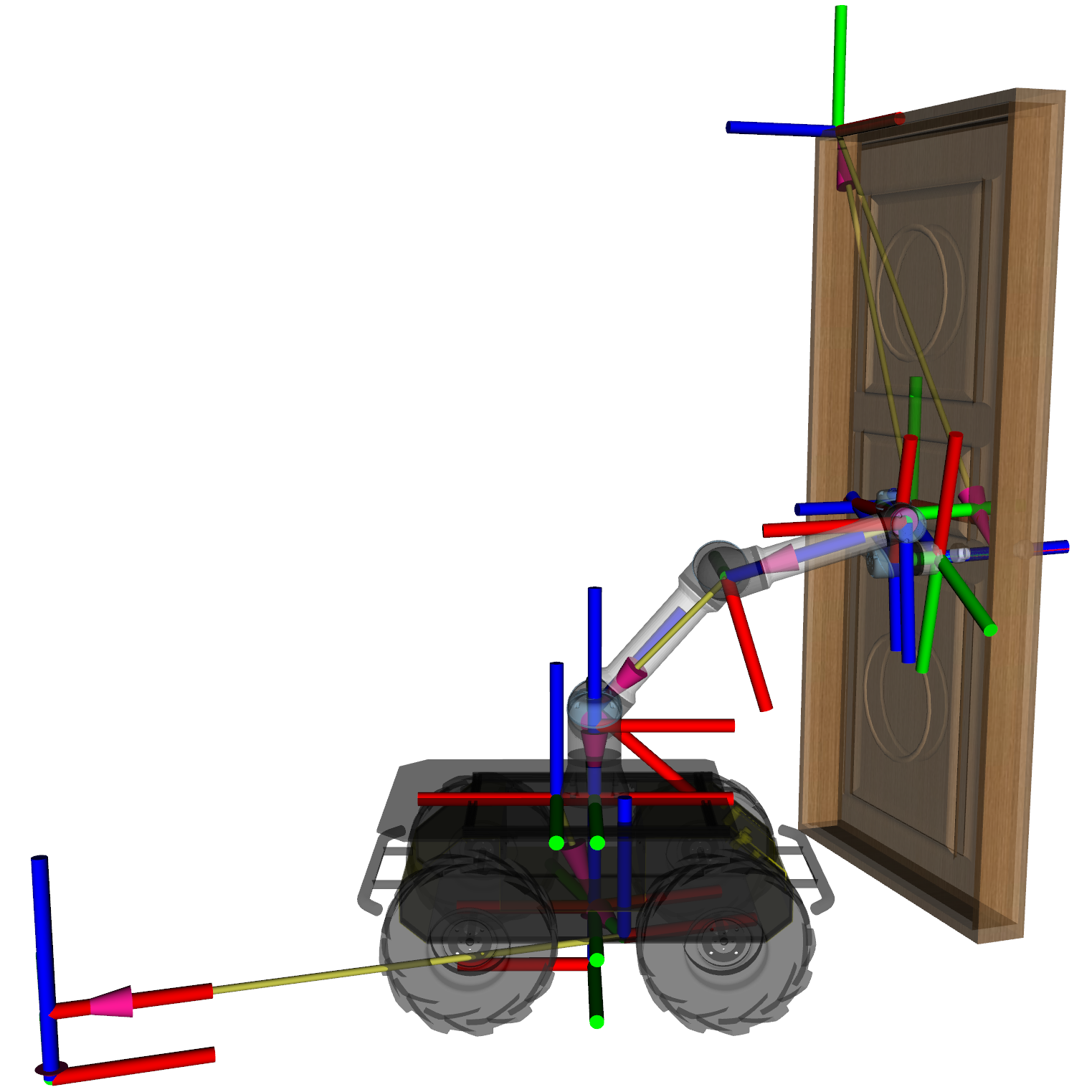}
        \caption{\ac{vkc}}
        \label{fig:vkc_construction_exp}
    \end{subfigure}%
    \caption{(a--c) The construction process of a \ac{vkc}. The green diamond denotes the \textit{root} frame of a kinematic structure, the yellow triangle denotes a \textit{terminal} frame (\eg, robot end-effector or object attachable frame), the blue circle denotes other frames, and the red trail in (c) highlights a constructed virtual kinematic chain. (d) The corresponding \ac{vkc} in opening the door.}
    \label{fig:vkc_construction}
\end{figure}

\section{\acf{vkc} Modeling}\label{sec:modeling}

The objective of the modeling is to construct a serial \ac{vkc} $\mathcal{C}^V$ by composing the kinematic of the mobile base, the arm, and the object to be manipulated.

\paragraph*{Original Structure}

The kinematic models of the robot and the object being manipulated are represented by two kinematic trees, $\mathcal{TG}^R$ and $\mathcal{TG}^O$, respectively. The highlighted area in \cref{fig:vkc_construction_a} indicates the original kinematic chain of interests: $\mathcal{C}^R$ for robots, and $\mathcal{C}^O$ for objects.

\paragraph*{Kinematic Inversion}

To construct a serial \ac{vkc} by inserting a virtual joint (\ie, an \textit{attachment}) between the robot's end-effector frame $\mathcal{F}^R_{ee}$ and the object's attachable frame $\mathcal{F}^O_{at}$, one has to invert the kinematic relationship of the object $\mathcal{C}^O$ as shown in the right panel of \cref{fig:vkc_construction_b}. Of note, such an inversion is not equivalent to simply reversing the parent-child relationship between the two adjacent links; the transformation between them must also be properly updated since a joint (\ie, revolute/prismatic) typically constrains the child link's motion \wrt \textit{child link's} frame.

The spatial relationships between any two frames can be represented by a homogeneous transformation.
A formal inversion of kinematic chains with multiple links is given by:
\begin{align}
    \small
    \label{eqn:gen_vkc_inv}
    {}^{\ \ \,i}_{i-1}{T}_\mathrm{inv} = {}_{i+1}^{\ \ \,i}{T}^{-1} = 
    \begin{bmatrix}
    {}_{i+1}^{\ \ \ i}{R^T} & -{}_{i+1}^{\ \ \ i}{R^T}~{}_{i+1}^{\ \ \ i}{p} \\
    \mathbf{0} & 1
    \end{bmatrix},
\end{align}
where ${}^a_bT$ is the transformation from the link $a$ frame to the link $b$ frame. Hence, ${}_{i+1}^{\ \ \,i}{T}$ is the transformation from the link $i$ frame to the link $i + 1$ frame in the original kinematic chain, whereas ${}^{\ \ \,i}_{i-1}{T}_\mathrm{inv}$ represents the transformation from the link $i$ frame to the link $i - 1$ frame in the inverted kinematic chain. \cref{eqn:gen_vkc_inv} describes the parent link of the link $i$ in the original kinematic chain becomes the child link of the link $i$ in the inverted kinematic chain, where the transformation of the inverted joint is given by ${}^{\ \ \,i}_{i-1}{T}_\mathrm{inv}$.



\paragraph*{\ac{vkc} Construction}

After inverting the object kinematic chain $\mathcal{C}^O$, a virtual kinematic chain can be constructed by adding a virtual joint (revolute, prismatic, or fixed) between $\mathcal{F}^{O}_{at}$ and $\mathcal{F}^{R}_{ee}$, whose transformation is denoted as ${}^{ee}_{at}{T}$. Next, a virtual base, whose frame is $\mathcal{F}^{V}_{b}$, is further inserted to enable a joint optimization of the locomotion and manipulation. Two perpendicular prismatic joints and a revolute joint are added between the virtual base and the robot base to imitate a planar motion between the mobile base and the ground, while ensuring the virtual kinematic chain ($\mathcal{C}^V$, highlighted by red in \cref{fig:vkc_construction_d}) remains serial.

\cref{fig:vkc_construction_exp} shows a constructed \ac{vkc} in opening a door. The above procedure produces a serial kinematic chain: Its base is a virtual fixed link in the environment, and its end-effector is at the link of the door that connects to the ground. The mobile base and the arm become the links embedded in the chain, and their poses during the door opening are calculated implicitly given the end-effector pose (\ie, the angle of the revolute hinge of the door). In other words, the final robot state is jointly optimized with trajectory, task goal, and kinematic constraints, without further efforts on finding the robot goal state or goal space for trajectory generation. This example demonstrates the possibilities to design robot actions and predicates without explicitly specifying the complete robot goal state (see \cref{sec:task}) while still being plausible at the motion level (see \cref{sec:motion}).

\section{Task Planning on \texorpdfstring{\ac{vkc}}{}}\label{sec:task}

Following the classic formalization of task planning, we describe the environment by a set of states $\mathcal{S}$. Possible transitions between these states are defined by $\mathcal{T}\subseteq\mathcal{S}\times\mathcal{S}$, where a transition $t=\langle s,s' \rangle \in\mathcal{T}$ alters the environment state from $s\in\mathcal{S}$ to $s'\in\mathcal{S}$. The goal of the task planning problem is to identify a sequence of transitions that alters the environment from its initial state $s_0\in S$ to a goal state $s_g\in S_g$, where $S_g\subseteq S$ is a set of goal states.

We primarily consider the task planning problems in mobile manipulation, which require the robot to account for its base, arm, and the object being interacted (\eg, pick and place, door/drawer opening). We formulate the task planning problems and implement the planning domains using \ac{pddl}.

In \ac{pddl}, the environment state $s$ is described by a set of \textit{predicates} that hold true. Specifically:
\setstretch{0.98}
\begin{itemize}[leftmargin=*,noitemsep,nolistsep,topsep=0pt]
    \item \texttt{(vkcState ?r ?q)}: A sub-chain \texttt{?r} (\eg, the base, an arm, or even a \ac{vkc}) of a \ac{vkc} is at configuration \texttt{?q} in joint space.
    \item \texttt{(objConf ?o ?s)}: An object \texttt{?o} is at the configuration \texttt{?s} in SE(3).
    \item \texttt{(free ?v)}: The robot end-effector is free to grasp.
    \item \texttt{(carry ?o ?v)}: The robot end-effector is carrying an object \texttt{?o}.
\end{itemize}
In this paper, to focus on demonstrating the benefit of task planning with \ac{vkc}, we pre-sampled feasible configurations \texttt{?s} for all objects and corresponding grasping poses.

Transitions in \ac{pddl} are modeled by \texttt{actions}. Each \texttt{action} takes parameters as input and can be called only when its preconditions hold true. After an \texttt{action} is called, its effect indicates how the states in the current environment change from preconditions. Thanks to the advantages introduced by the \ac{vkc}, three simple action definitions---\texttt{goto-vkc}, \texttt{pick-vkc}, and \texttt{place-vkc}---are sufficient to handle various mobile manipulation tasks, from pick-and-place in different setups to foot-arm coordinated and constrained motions (\eg, door/drawer opening). Below is an example of the definitions for three actions; see \cref{fig:exp1-b,fig:exp1-c,sec:exp1} for a comparison between \ac{vkc}-based \ac{pddl} and a standard \ac{pddl} for mobile manipulators.

{\small
\begin{verbatim}
(:action goto-vkc
   :parameters (?r ?from ?to)
   :precondition (vkcState ?r ?from)
   :effect (and (vkcState ?r ?to)
                (not (vkcState ?r ?from))))
                
(:action pick-vkc
   :parameters (?o - obj ?s - state ?v - vkc)
   :precondition (and (objConf ?o ?s)
                      (free ?v))
   :effect  (and (carry ?o ?v)
	            (not (objConf ?o ?s))
	            (not (free ?v))))
	            
(:action place-vkc
   :parameters (?o - obj ?s - state ?v - vkc)
   :precondition (and (carry ?o ?v)
                      (not (occupied ?s)))
   :effect  (and (not (carry ?o ?v))
                      (objConf ?o ?s)
                      (free ?v)))
\end{verbatim}}

\section{\texorpdfstring{\ac{vkc}}{} Facilitates Motion Planning}\label{sec:motion}

The conventional task planning setup usually assumes a robot already knows how to execute the actions defined in the task domain and, therefore, does not generate actionable motion trajectories for the robot. However, in practice, this assumption does not always hold as many abstract actions defined in the task domain are difficult to be instantiated at the motion level. This section discusses how the \texttt{actions} defined using \ac{vkc} can properly form a motion planning problem solvable by existing motion planners.

\subsection{From Task to Motion}

We start by making the connections between the action semantics and the actual manipulation behaviors, followed by explaining how the predicates and variables in the action definitions are processed by motion planners.

\setstretch{0.99}

\paragraph*{\texttt{goto-vkc} (\texttt{r}, $\mathbf{q_1}$, $\mathbf{q_2}$)}

This predicate moves the \ac{vkc} from the current pose $q_1$ to a desired pose $q_2$ for a chain \texttt{r}. It represents the tasks that do not require interaction with the environment, wherein the \ac{vkc} structure remains unchanged. Pure navigation is a typical action falling into this category. For example, \texttt{goto-vkc} (\texttt{base}, $q^b_1$, $q^b_2$) moves the robot to the location specified in $q^b_2$. Another example is to manipulate a picked object from the current pose $q_1$ to a certain pose $q_2$, \ie, \texttt{goto-vkc} (\texttt{vkc}, $q_1$, $q_2$)

\paragraph*{\texttt{pick-vkc} (\texttt{object}, $\mathbf{s}$, \texttt{vkc})}

This predicate moves the \ac{vkc} to the \texttt{object} to be manipulated and extends the current \ac{vkc} structure by adding a virtual joint to connect the \texttt{object} and the arm's end-effector at state $s$. Here, the state could be interpreted as a grasping pose, the transformation between the robot gripper and the object to be manipulated (\ie, ${}^{ee}_{at}{T}$). \texttt{pick-vkc} represents the group of tasks that require mobile manipulators to interact with the environment, \eg, picking up an object or grasping a handle.

\paragraph*{\texttt{place-vkc} (\texttt{object}, $\mathbf{s}$, \texttt{vkc})}

This predicate moves the \texttt{object} connected to \texttt{vkc} to a goal pose $s$, while the object to be manipulated is incorporated into the \ac{vkc} and imposes kinematic constraints to the planner. Once reaching the goal pose, \texttt{place-vkc} breaks the current \ac{vkc} at the virtual joint where it connects the mobile manipulator and the \texttt{object}, and the \texttt{object} will be placed at where it was disconnected from \ac{vkc}. \texttt{place-vkc} represents the group of tasks that mobile manipulators stop interacting with the environment, such as placing an object on the table.

In motion planning, configuration space $Q$ describes the environment state. $Q$'s dimension $n$ equals to \ac{vkc}s' degrees of freedom. A collision-free subspace $Q_{\text{free}}\subseteq Q$ is the space that \ac{vkc}s can traverse freely without colliding with the environment or itself. The problem of motion planning on \ac{vkc} is equivalent to finding a collision-free path $\mathbf{q}_{1:T}\in Q_{\text{free}}$ from the initial pose $\mathbf{q}_{1}\in Q_{\text{free}}$ to reach the final state $\mathbf{q}_T\in Q_{\text{free}}$. Each action predicate requires to form a motion planning problem due to the kinematic structure changes.

\subsection{Optimization-based Motion Planning}\label{sec:opt_motion}

Finding a collision-free path $\mathbf{q}_{1:T}\in Q_{\text{free}}$ for given tasks can be formulated by trajectory optimization, \eg, CHOMP~\cite{ratliff2009chomp} and TrajOpt~\cite{schulman2014motion}. The objective function of the trajectory optimization can be formally expressed as:
\begin{equation}
    \small
    \underset{\mathbf{q}_{1:T}}{\min} \sum_{t=1}^{T-1} || W_{\text{vel}}^{1/2} \;\; \delta\mathbf{q}_{t} ||_2^2
    \; + \sum_{t=2}^{T-1} || W_{\text{acc}}^{1/2} \;\; \delta\dot{\mathbf{q}}_{t} ||_2^2,
    \label{eqn:objective}
\end{equation}
where $\mathbf{q}_{1:T}$ is the trajectory sequence $\{\mathbf{q}_1, \mathbf{q}_2, \ldots, \mathbf{q}_T\}$, and $\mathbf{q}_{t}$ the \ac{vkc} state at the $t^\mathrm{th}$ time step. We penalize the overall velocities and acceleration of every joint with diagonal weights $W_{\text{vel}}$ and $W_{\text{acc}}$ for each joint, respectively. 

Meanwhile, the constructed \ac{vkc} should also be subject to kinematic constraints of the robot and the environment: 
\begin{equation}
    \small
    h_{\text{chain}}(\mathbf{q}_t) = \mathbf{0}, \; \forall t = 1, 2, \ldots, T, \label{eqn:chain_cnt}
\end{equation}
including forward kinematics and closed chain constraints. We can formulate the task goal as an inequality constraint:
\begin{equation}
    \small
    || f_{\text{task}}(\mathbf{q}_T) - \mathbf{g} ||^2_2 \leq \xi_{\text{goal}}, \label{eqn:goal_cnt}
\end{equation}
which bounds the element-wise squared $\ell^2$ norm between the final state in the goal space $f_{\text{\text{task}}}(\mathbf{q}_T)$ and the task goal $\mathbf{g}$ $\in\mathbb{R}^k$\ ($k\leq n$) with a tolerance $\xi_{\text{goal}}$. The function $f_{\text{task}}: Q\rightarrow\mathbb{R}^k$ is a task-dependent function that maps the joint space of a \ac{vkc} to the goal space that differs from task to task. This definition relaxes hard constraints of goal state and optimized the other $n-k$ states with objective function \cref{eqn:objective}. Of note, \cref{eqn:chain_cnt,eqn:goal_cnt} are not the only forms of constraints that a \ac{vkc}-based approach can incorporate; in fact, it is straightforward to add additional task constraints to the same optimization problem in \cref{eqn:objective}, depending on various task-specific requirements.
In this paper, we further impose several additional safety constraints, including joint limits, bounds for joint velocity and acceleration, and link-link and link-object collisions. A more detailed analysis of the optimization framework can be found in Jiao \etal~\cite{jiao2021consolidated}.

\subsection{Sampling-based Motion Planning}\label{sec:sample_motion}

Alternatively, motion planning on \ac{vkc} can also be viewed as a search procedure in the configuration space $\mathcal{Q}_{\text{free}}$. Given a path planning problem within $\mathcal{Q}_{\text{free}}$, a sampling-based method would attempt to find a set of collision-free way points that start from an initial configuration $\mathbf{q}_0 \in \mathcal{Q}_{\text{free}} $ and end in the goal configuration $\mathbf{q}_\text{{goal}}$ $\in \mathcal{Q}_{\text{free}} $. 

\ac{rrt} is a probabilistically complete search algorithm that incrementally expands a collection of directional nodes $ \mathcal{T}$ to explore space~\cite{lavalle2000rapidly}. In this paper, we adopted a \ac{rrt}-connect algorithm~\cite{kuffner2000rrt} from the \ac{ompl}~\cite{sucan2012open} as our sampling-based motion planner, which initiates exploration from $\mathbf{q}_0$ and $\mathbf{q}_{goal}$ concurrently. 

Unlike the optimization-based method mentioned in \cref{sec:opt_motion}, way-points collected by \ac{rrt}-connect are not smoothed by an objective function during search; instead, interpolation was performed after the search is complete for a smooth trajectory to be executed on a mobile manipulator.

\begin{figure}[t!]
    \centering
    \includegraphics[width=0.88\linewidth]{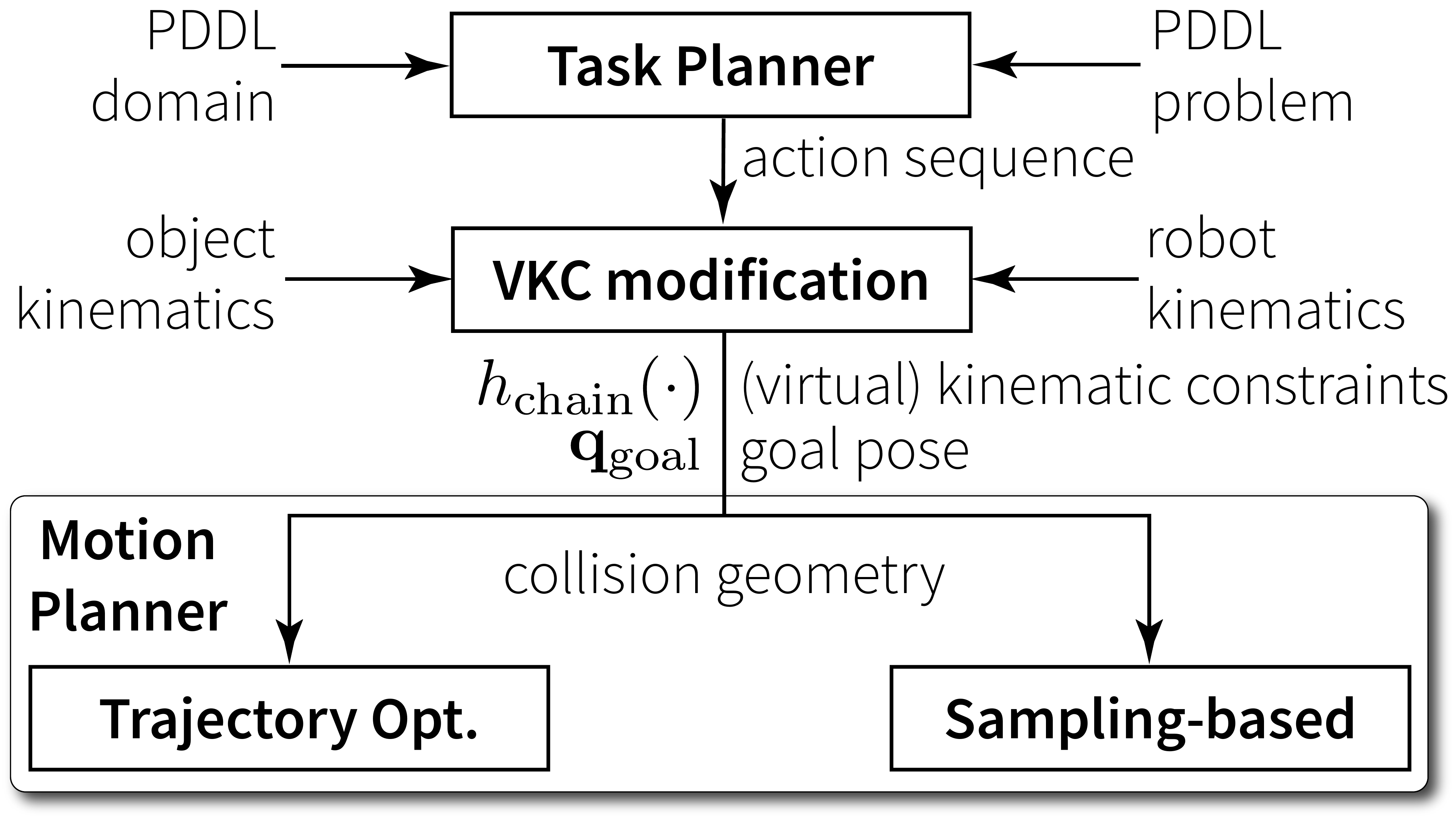}
    \caption{\textbf{The computing logic of instantiating the \texttt{actions} in a task plan to trajectories at the motion level.} Each action symbol encodes a (virtual) kinematic chain and a goal pose, which are sufficient for a motion planner given the environmental constraints.}
    \label{fig:figure3}
\end{figure}

\cref{fig:figure3} summarizes the computing logic of instantiating the \texttt{actions} to motion trajectories. The action sequence produced by the task planner encodes how the \ac{vkc} changes over each action and its desired goal pose. Together with environmental constraints (\eg, the actual robot kinematics and the objects' geometry), the information provided by the \ac{vkc}-based task planner is sufficient for a typical motion planner to produce a feasible trajectory from $\mathbf{q}_0$ to $\mathbf{q}_{\text{goal}}$.

\setstretch{0.945}

\section{Experiment}\label{sec:exp}

We conduct a series of experiments to evaluate the efficacy of the proposed \ac{vkc} perspective for planning mobile manipulation tasks in simulations. The first experiment compares the designs of \ac{pddl} definition with \ac{vkc} or without \ac{vkc} and their corresponding planning efficiency. Since the action definitions can be arbitrarily abstract at the symbolic task level, we further validate the \ac{vkc}-based action design in the second experiment that it indeed provides sufficient information for motion planners to produce feasible trajectories. Finally, in the third experiment, we showcase how the \ac{vkc} perspective empowers more complex task planning.

\subsection{Simplifying Task Domain}\label{sec:exp1}

\begin{figure}[t!]
    \centering
    \begin{subfigure}[b]{0.5\linewidth}
        \centering
        \includegraphics[width=0.9\linewidth]{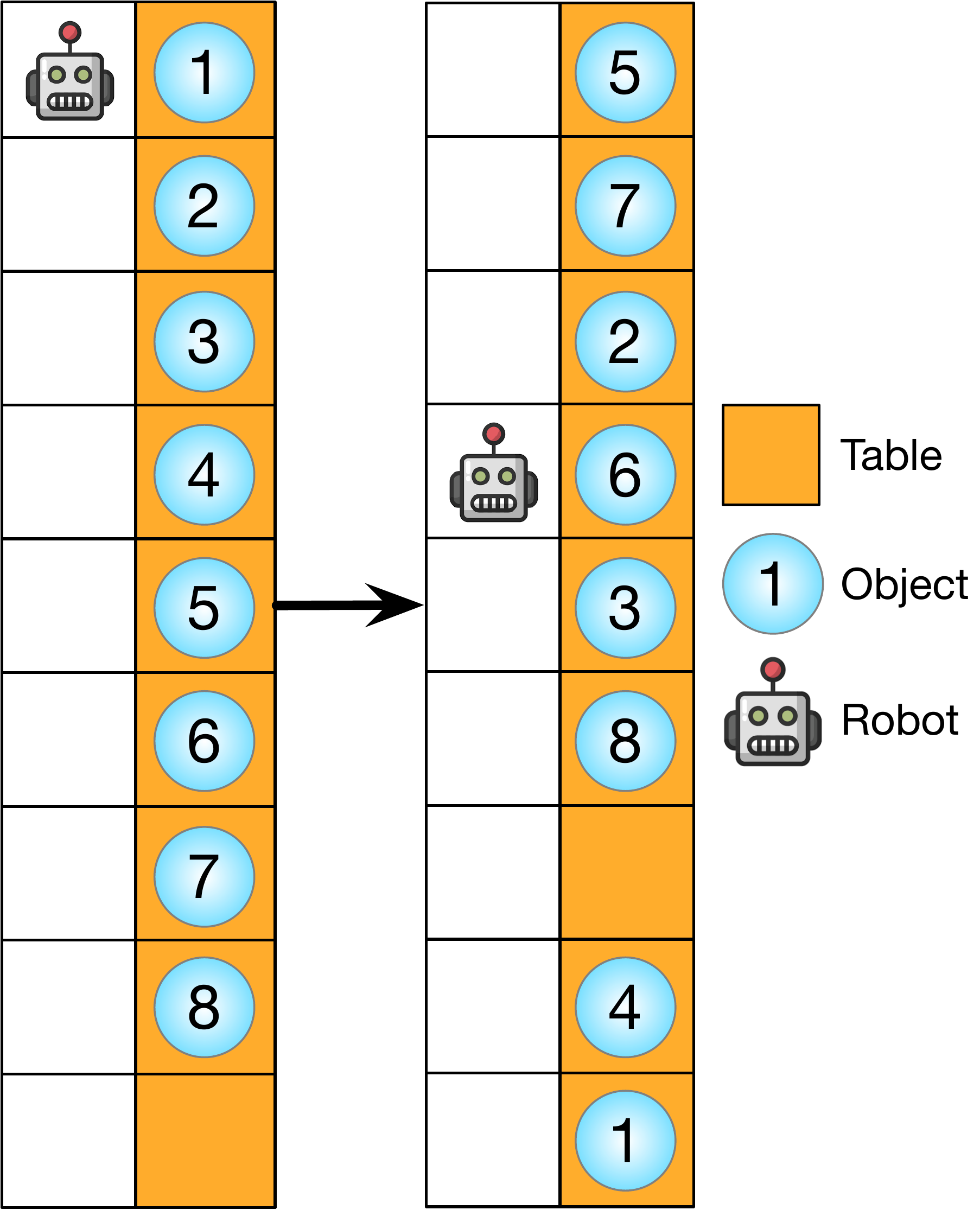}
        \caption{Experimental Setup}
        \label{fig:exp1-a}
        \vspace{3pt}
        \includegraphics[width=0.98\linewidth]{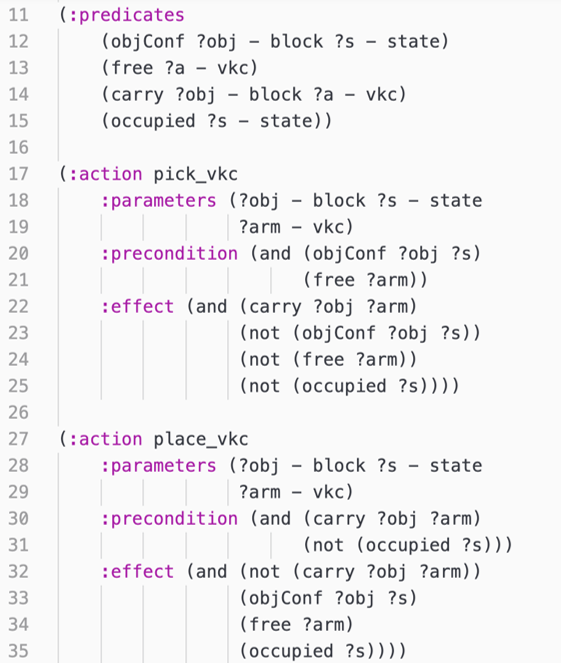}
        \caption{\ac{vkc}-based \ac{pddl}}
        \label{fig:exp1-b}
    \end{subfigure}%
    \begin{subfigure}[b]{0.5\linewidth}
        \centering
        \includegraphics[width=0.98\linewidth]{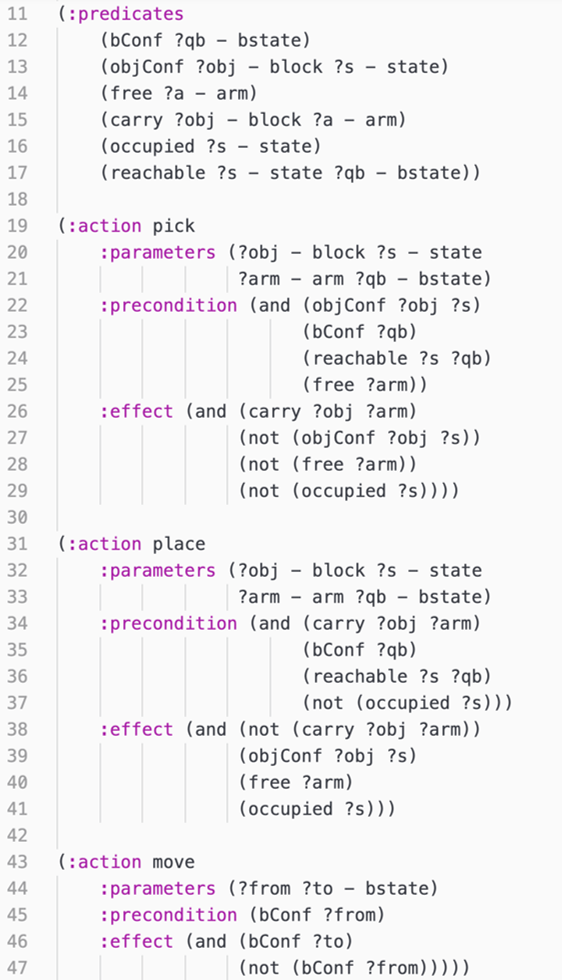}
        \caption{Conventional \ac{pddl}}
        \label{fig:exp1-c}
        \includegraphics[width=0.98\linewidth,trim={0cm 0.2cm 1cm 0.7cm},clip]{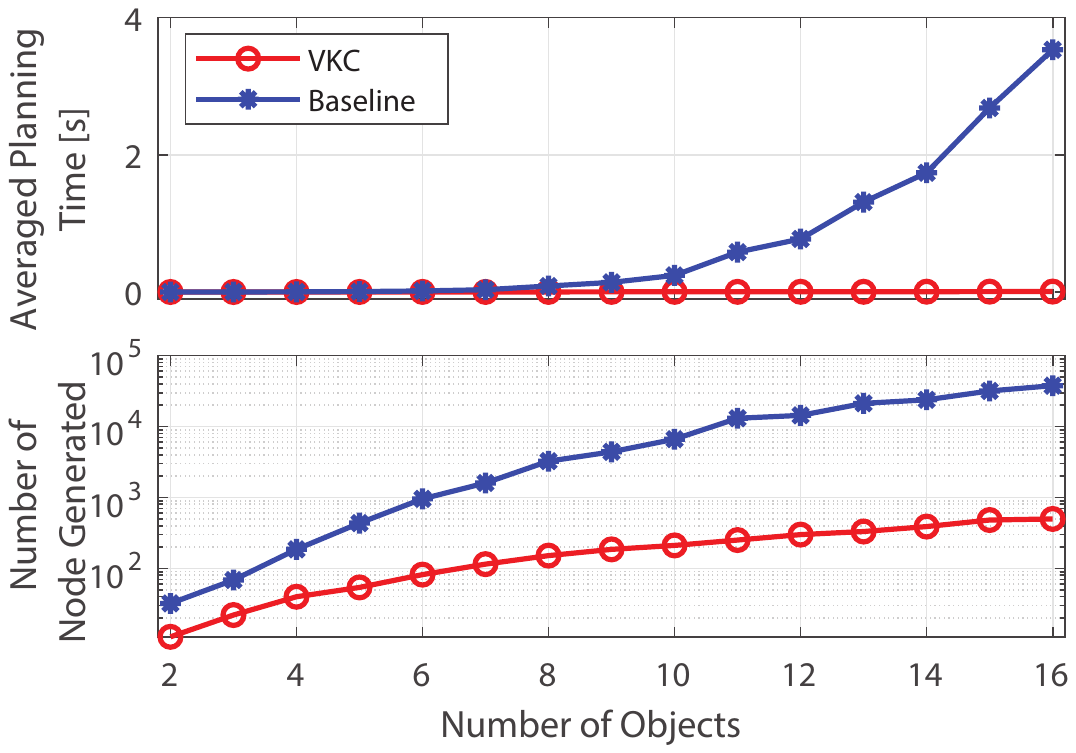}%
        \caption{Performance Comparison}
        \label{fig:exp1-d}
    \end{subfigure}%
    \caption{\textbf{\ac{vkc}-based domain specification improves the task planning efficacy.} (a) An example setup of re-arranging 8 objects on 9 tables; one table can only support one object. (b) The \ac{vkc}-based \ac{pddl} specification has less variables and more abstract actions than (c) a conventional \ac{pddl} specification. (d) The \ac{vkc}-based domain specification allows a solver to search for a feasible plan for tasks of re-arranging 2 to 16 objects with significantly less time and generated nodes in search (\ie, less memory).}
    \label{fig:exp1}
\end{figure}

Since the \ac{vkc} perspective treats the base, the arm, and the object to be manipulated as a whole, designing the planning domain becomes much simpler. In this experiment, we focus on an object-arrangement task, where the robot is tasked to re-arrange $m$ objects on $m+1$ tables into the desired order while satisfying the constraint that each table can only support one object. \cref{fig:exp1-a} shows a typical example of this task's initial and goal configuration with $m=8$ objects, randomly sampled in each experimental trial. 

\cref{fig:exp1-b} shows a \ac{pddl} domain designed by the actions mentioned in \cref{sec:task}, which requires less predicates and provides more abstract actions compared with those designed by conventional domain definition shown in \cref{fig:exp1-c}. Specifically, the conventional method would require (i) more predicates to describe the mobile base's states and thus more complex preconditions for actions, (ii) one more action to control the mobile base, and (iii) more parameters for other actions. To solve for a task plan, we adopt the \ac{iws} algorithm~\cite{lipovetzky2014width}; it is a width-limited version of the \ac{bfs} that repeatedly runs with increasing width limits until a feasible task plan is found. If no feasible task plan could be found within the maximum width limit of the \ac{iws}, a traditional \ac{bfs} with no width limit will be deployed to search for a solution. 

In experiments, we run 50 trials for each setup; see the result summary in \cref{fig:exp1-d}. As the task complexity increases, the average planning time and the number of nodes generated in search (\ie, memory required) increase relatively slowly for the \ac{vkc}-based task plan. In comparison, the baseline using conventional methods increases much more rapidly.

This result is evident. As we can see in \cref{fig:exp1-d}, planning in the non-\ac{vkc} version of the task domain requires exploring more nodes at each depth level to find a plausible pose for the mobile base. It also requires more actions to accomplish the task, which further yields a deeper depth during the search. Suppose there are $N$ nodes on average to be generated at each depth level of the search algorithm, and a feasible solution is found at depth $d$, the total number of nodes being generated is $N^d$. In theory, when the search algorithm performs in the \ac{vkc} domain, the total number of generated node is $(c_1N)^{c_2d}$, where $c_1 \leq 1, c_2 \leq 1$. In the task with 16 objects, our experiment empirically finds $c_1 = 0.75$ and $c_2 = 0.22$ on average over 50 trails.

Taken together, the results in the first experiment demonstrated that \ac{vkc}-based task planning requires much fewer explorations in both width and depth during the search algorithm, therefore achieving higher efficacy with less memory.

\subsection{Improving Mobile Manipulation}

In general, actions that are more abstract and with fewer variables in the planning domain specification would lead to more efficient task planning, but simultaneously could result in less success rate in generating feasible plans at the motion level. In this experiment, we validate that the \ac{vkc}-based task planning provides efficacy at the task level and maintains a high success rate at the motion level. Based on the generated task plans (\ie, action sequences) and the encoded information (as described in \cref{sec:motion}), we apply a trajectory optimization-based motion planner and a sampling-based motion planner and evaluate how well they can produce feasible motion trajectories for the given task.

\begin{figure}[t!]
    \centering
    \begin{subfigure}[c]{\linewidth}
        \centering
        \includegraphics[width=\linewidth]{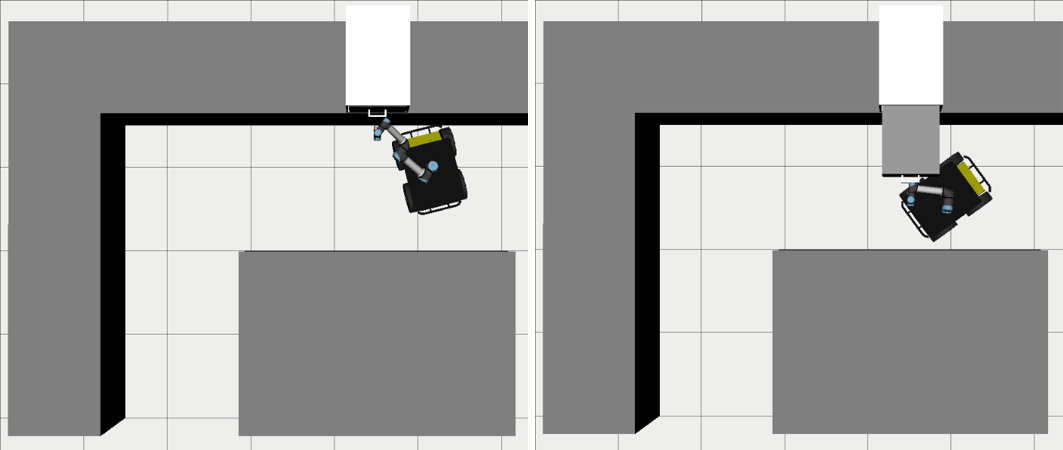}
        \caption{The drawer opening task.}
        \label{fig:exp2_task}
    \end{subfigure}%
    \\
    \begin{subfigure}[c]{\linewidth}
        \centering
        \includegraphics[width=\linewidth]{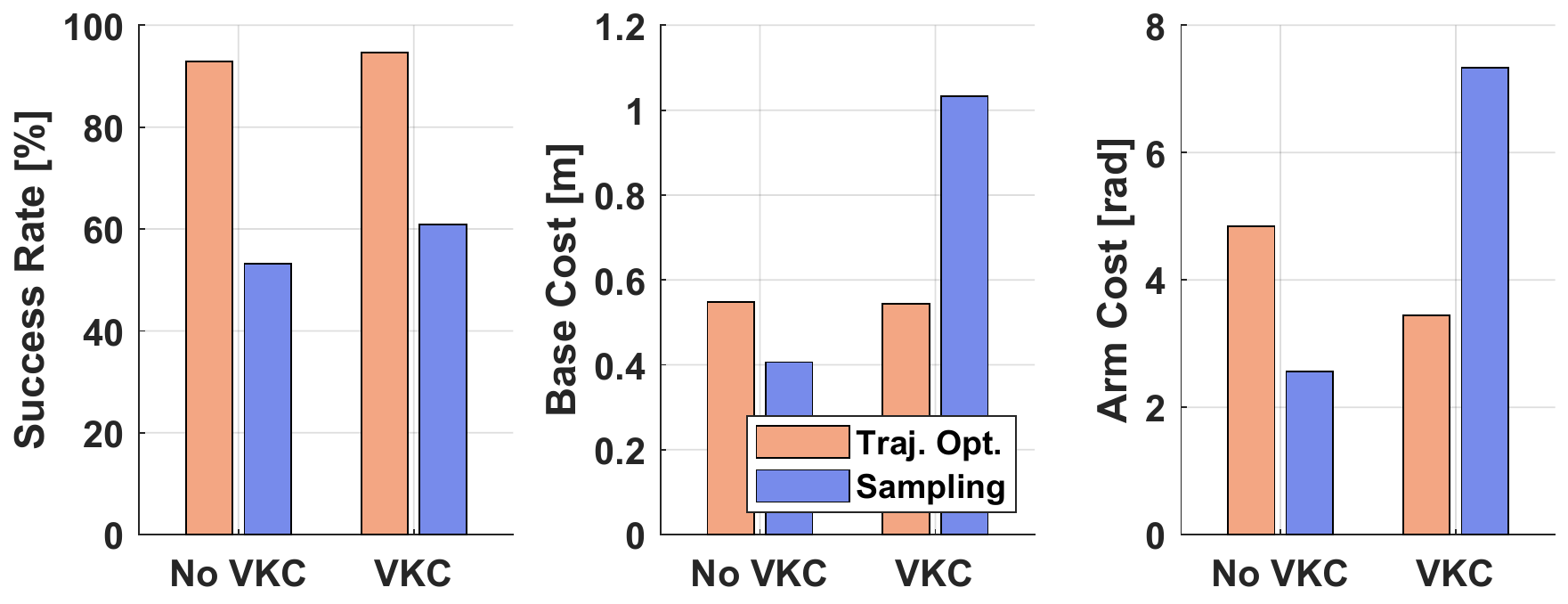}
        \caption{Success rates and the corresponding base and arm movements.}
        \label{fig:exp2_result}
    \end{subfigure}%
    \caption{\textbf{Instantiating the task plans to motions} in (a) a drawer opening task. The domains, one with \ac{vkc} and the other without, are specified similar to \cref{fig:exp1-b,fig:exp1-c}. The generated task plans are processed by an optimization-based and a sampling-based motion planner. (b) Task success rates, and base and arm costs. Failure cases for sampling include time-out for both sub-tasks: 5 mins for reach, 50 seconds for open}
    \label{fig:exp2}
\end{figure}

Specifically, we consider the task of pulling opening a drawer; see \cref{fig:exp2_task}. The task plans: (i) \texttt{place-vkc} (\texttt{drawer}, $s^1_{\texttt{d}}$, \texttt{vkc}), (ii) \texttt{move} ($q^0_{\texttt{b}}$, $q^1_{\texttt{b}}$) + \texttt{place} (\texttt{drawer}, $s^1_{\texttt{d}}, \texttt{arm}$, $q^1_{\texttt{b}}$), are produced by two \ac{pddl}s specified with and without \ac{vkc}, respectively. We compare the success rate of executing the trajectories planned by trajectory optimization and sampling motion planning methods described in \cref{sec:opt_motion,sec:sample_motion}, as well as the base and arm cost measured by the distances they travel; see \cref{fig:exp2_result}. 

\setstretch{1}

The trajectory optimization-based motion planner can produce feasible trajectories for the given task with high success rates; the produced trajectories are more efficient in terms of shorter base and arm traveling distances. Typically, sampling-based motion planners would struggle in incorporating kinematic and safety constraints due to naturally unconstrained configuration spaces, which need extra effort to accommodate extra kinematic constraints~\cite{kingston2019exploring}, making it less suitable for such tasks. However, it is still more successful in producing feasible trajectories under the \ac{vkc} specification compared with the setting without \ac{vkc}. The most significant drawbacks of sampling-based motion planners are the high execution costs and violation of safety limits.

\subsection{Solving Tasks with Multiple Steps}

Complex multi-step mobile manipulation tasks with long action sequences can also be easily accomplished using the action set introduced by the \ac{vkc}-based task planner described in \cref{sec:task}. These actions contain high-level task semantics that could be adapted to various tasks; \eg, attaching to the doorknob could be expressed by a $\texttt{pick-vkc}$ action, and open the door to a certain angle could be expressed by a $\texttt{place-vkc}$.

\begin{figure*}[t!]
    \centering
    \begin{subfigure}[c]{\linewidth}
        \centering
        \includegraphics[width=\linewidth,trim={0cm 0.35cm 0cm 0cm},clip]{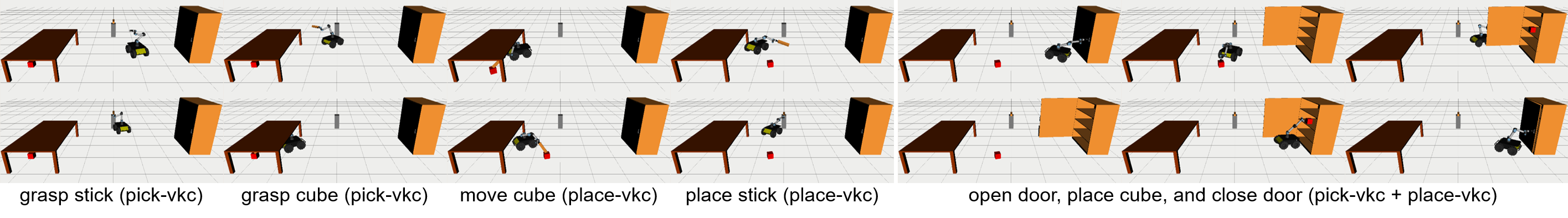}
        \caption{}
        \label{fig:exp3_a}
    \end{subfigure}%
    \\%
    \begin{subfigure}[c]{0.22\linewidth}
        \centering
        \includegraphics[width=\linewidth]{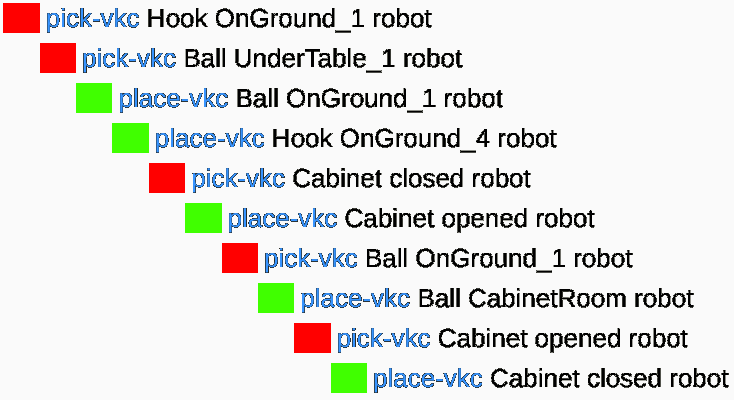}
        \caption{}
        \label{fig:exp3_b}
    \end{subfigure}%
    \begin{subfigure}[c]{0.78\linewidth}
        \centering
        \begin{overpic}[width=\linewidth,trim={2.8cm 0cm 2cm 0cm},clip]{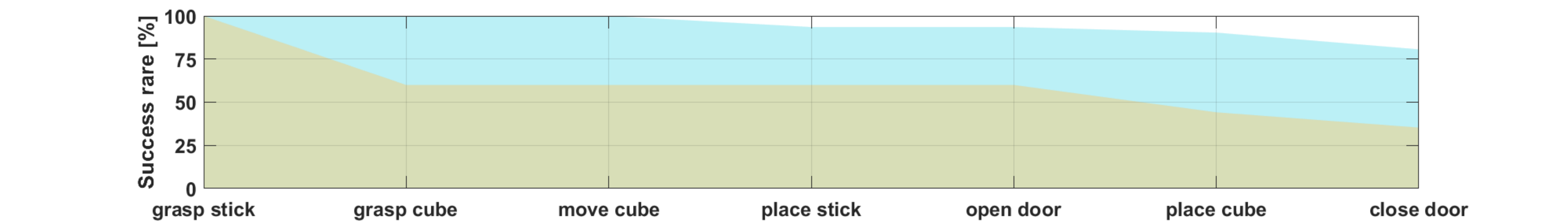}
        \put(20, 12){\color{black} with \ac{vkc}}
        \put(20, 5){\color{black} without \ac{vkc}}
        \end{overpic}
        \caption{}
        \label{fig:exp3_c}
    \end{subfigure}%
    \caption{(a) The \ac{vkc}-based task planner can easily scale up to a complex multi-step task, which can be (b) succinctly expressed by merely two actions defined based on the \ac{vkc}s. (c) More abstract action definitions introduced by \ac{vkc} instantiate better at the motion level, possessing an excellent foot-arm coordination in each step of the task. Without \ac{vkc}, to ensure successful planning for tasks that require foot-arm coordination, several actions must be executed together to complete certain steps in the task.}
    \label{fig:exp3}
\end{figure*}

\cref{fig:exp3_a} qualitatively shows a complex multi-step task planning using the \ac{vkc}-based domain specification and instantiating that to motions. For a more fair comparison, in addition to the initial and goal state of the environment, both the \ac{vkc} and non-\ac{vkc} methods are provided with the (identical) grasping poses for all movable objects, but not the corresponding robot state. In this task, a mobile manipulator needs to (i) grasp the stick, (ii) fetch the cube under the table using the stick, which is otherwise challenging to reach, (iii) move the cube outside, (iv) place the stick down, (v) grasp the cabinet and open it, (vi) place the cube inside the cabinet, and (vii) close the cabinet door. At each trial, the mobile manipulator is randomly placed in the environment.

\cref{fig:exp3_b} illustrates that the above complex multi-step task can be accomplished by using only two abstract actions defined based on \ac{vkc}, one action in each step. Without the \ac{vkc} perspective, significantly more effects must first be devoted to designing the planning domain. Furthermore, to ensure successful planning of actions that require foot-arm coordination, each step may require several actions to be executed together; see \cref{tab:exp3} for a comparison between the two setups. Even after the additional efforts of specifying base pose from a feasible region, its accumulated success rate at the motion level produced by the corresponding actions still underperforms the \ac{vkc} version, shown in \cref{fig:exp3_c}. Without \ac{vkc}, the motion planner particularly suffers at step 2 when the robot needs to fetch the cube in a confined space, as it requires the planner to deliver proper navigation and manipulation with excellent foot-arm coordination (\ie, coordinating \texttt{move} and \texttt{place}). In sum, this experiment demonstrates that \ac{vkc}-based mobile task planning for mobile manipulation tasks is advantageous by simplifying domain specification and improving motion planning.
 
\begin{table}[ht!]
    \centering
    \caption{\textbf{Actions and predicates in the defined planning domains.} Without \ac{vkc}, more actions must be specified, and extra predicates are required for generating a feasible task plan.}
    \label{tab:exp3}
    \resizebox{\columnwidth}{!}{%
        \begin{tabular}{@{\hskip0pt}c@{\hskip0pt}c@{\hskip6pt}c@{\hskip6pt}l}
            \hline
            \textbf{Setup} & \textbf{Group} & \textbf{Notation} & \textbf{\quad{}\quad{}\quad{}\quad{}\quad{}\quad{}Description}\\ \hline \hline
            \multirow{9}{*}{\rotatebox[origin=c]{90}{\ac{vkc}}} &
            \multirow{2}{*}{Actions} & \texttt{pick-vkc$(o,s,v)$} & \multirow{2}{*}{see \cref{sec:task}}\\ 
            & & \texttt{place-vkc$(o,s,v)$} & \\
               \cline{2-4}
            & \multirow{7}{*}{Predicates}
                 & \texttt{Graspable$(o, v)$} & Check if robot $v$ is able to grasp object $o$ $v$ \\ 
                & & \texttt{RigidObj$(o)$} & Check if object $o$ is rigid object \\
                & & \texttt{ArtiObj$(o)$} & Check if object $o$ is articulated object \\
                & & \texttt{ToolObj$(o)$} & Check if object $o$ could be used as a tool \\ 
                & & \texttt{Occupied$(s)$} & Check if a position $s$ being occupied \\
                & & \texttt{Carried$(o)$} & Check if an object $o$ is carried by robot \\
                & & \texttt{ContainSpace$(o, s)$} & Check if a position $s$ being contained in the object $o$ \\\hline \hline
            \multirow{7}{*}{\rotatebox[origin=c]{90}{Non-\ac{vkc}}} & \multirow{5}{*}{Actions} 
            & \texttt{move$(s_1, s_2)$} & Move robot from $s_1$ to $s_2$ \\
            & & \texttt{pick$(o, s_1, s_2)$} & Pick the object $o$ at location $s_1$ given robot state $s_2$ \\
            & & \texttt{place$(o, s_1, s_2)$} & Place the object $o$ at location $s_1$ given robot state $s_2$ \\
            & & \texttt{open$(o, s_1, s_2)$} & Open the object $o$ at location $s_1$ given robot state $s_2$ \\
            & & \texttt{close$(o, s_1, s_2)$} & Close the object $o$ at location $s_1$ given robot state $s_2$ \\
            \cline{2-4}
            & Extra & \texttt{HasTool$(s)$} & Check if the robot at current state $s$ holding a tool \\
            & Predicates & \texttt{AbleToPick$(s)$} & Check if the robot at state $s$ is able to do pick action \\
            & for Non-VKC & \texttt{Reachable$(o, s)$} & Check if object $o$ is reachable by mobile base at state $s$ \\ 
            \hline
        \end{tabular}%
    }%
\end{table}

\section{Discussion and Conclusion}\label{sec:conclusion}

We present a \ac{vkc} perspective that improves the domain specifications for mobile manipulation task planning. By integrating the kinematics of the mobile base, the arm, and the object to be manipulated into a single \ac{vkc}, more abstract action symbols become possible and fewer predicates/variables/intermediate goals are required in designing the planning domain. In a series of experiments, we demonstrate that the \ac{vkc}-based domain specification using \ac{pddl} supports more efficient task planning, works better with existing motion planners, and scales up to more complex tasks compared with the one without \ac{vkc}. We argue the proposed \ac{vkc} perspective has significant potential in promoting mobile manipulation in real-world daily tasks. Our future work will explore \ac{tamp} on \ac{vkc}.

\paragraph*{Acknowledgement}
We thank Jiang Xin of the UCLA ECE Department for discussions on trajectory optimization.

\setstretch{0.93}
\balance
\bibliographystyle{ieeetr}
\bibliography{IEEEfull}

\end{document}